\documentclass[10pt,twocolumn,letterpaper]{article}

\usepackage{cvpr}              
\usepackage{amssymb} 
\usepackage{amsmath}
\usepackage{graphicx} 
\usepackage{booktabs}
\usepackage{multirow}

\definecolor{cvprblue}{rgb}{0.21,0.49,0.74}
\usepackage[pagebackref,breaklinks,colorlinks,allcolors=cvprblue]{hyperref}

\def\paperID{15166} 
\def\confName{CVPR}
\def\confYear{2026}

\title{Structure-Aware Fine-Grained Gaussian Splatting for \\Expressive Avatar Reconstruction}

\author{
	Yuze Su\textsuperscript{1}, \; Hongsong Wang\textsuperscript{2,3}, \; Jie Gui\textsuperscript{1,4,5}, \; Liang Wang\textsuperscript{6,7,8}\\
	$^{1}$School of Cyber Science and Engineering, Southeast University, Nanjing 210096, China\\
	$^{2}$School of Computer Science and Engineering, Southeast University, Nanjing 210096, China \\
	$^{3}$Key Laboratory of New Generation Artificial Intelligence Technology and Its Interdisciplinary \\
	Applications (Southeast University), Ministry of Education, China \\ 
	$^{4}$Purple Mountain Laboratories, Nanjing 210000, China \\
	$^{5}$Engineering Research Center of Blockchain Application, Supervision And Management\\ (Southeast University), Ministry of Education, China \\
	$^{6}$State Key Laboratory of Multimodal Artificial Intelligence Systems (MAIS), Beijing, China\\
	$^{7}$Institute of Automation, Chinese Academy of Sciences (CASIA), Beijing, China\\
	$^{8}$School of Artificial Intelligence, University of Chinese Academy of Sciences, Beijing, China \\
	\tt\small\{220245811, hongsongwang, guijie\}@seu.edu.cn;wangliang@nlpr.ia.ac.cn \\
}

\begin{document}
	\maketitle
	
	\begin{abstract}
		Reconstructing photorealistic and topology-aware human avatars from monocular videos remains a significant challenge in the fields of computer vision and graphics. While existing 3D human avatar modeling approaches can effectively capture body motion, they often fail to accurately model fine details such as hand movements and facial expressions. To address this, we propose Structure-aware Fine-grained Gaussian Splatting (SFGS), a novel method for reconstructing expressive and coherent full-body 3D human avatars from a monocular video sequence. The SFGS use both spatial-only triplane and time-aware hexplane to capture dynamic features across consecutive frames. A structure-aware gaussian module is designed to capture pose-dependent details in a spatially coherent manner and improve pose and texture expression. To better model hand deformations, we also propose a residual refinement module based on fine-grained hand reconstruction. Our method requires only a single-stage training and outperforms state-of-the-art baselines in both quantitative and qualitative evaluations, generating high-fidelity avatars with natural motion and fine details. The code is on Github: \url{https://github.com/Su245811YZ/SFGS}
	\end{abstract}
	
	\section{Introduction}
	Driven by the booming market of VR/AR and telepresence, photo-realistic digital humans have become a cornerstone for interactive applications~\cite{vid2avatar,xu2023vr}. Reconstructing photorealistic human avatars from monocular RGB videos has received growing attention in computer vision and graphics.
	
	Early personalized 3D human avatar generation techniques~\cite{3DGS,NeuMan,X-avatar} typically rely on parametric body models, such as SMPL~\cite{SMPL} and SPML-X~\cite{SMPLX2019}, to guide the reconstruction of high-fidelity geometry and appearance. However, these methods are constrained by the meshes obtained from scanning, which often struggle with modeling loose clothing, complex hairstyles, and posture-dependent wrinkles, leading to artifacts or misalignment issues due to inherent limitations in the mesh data.
	
	\begin{figure}[t]
		\centering
		\includegraphics[width=\linewidth]{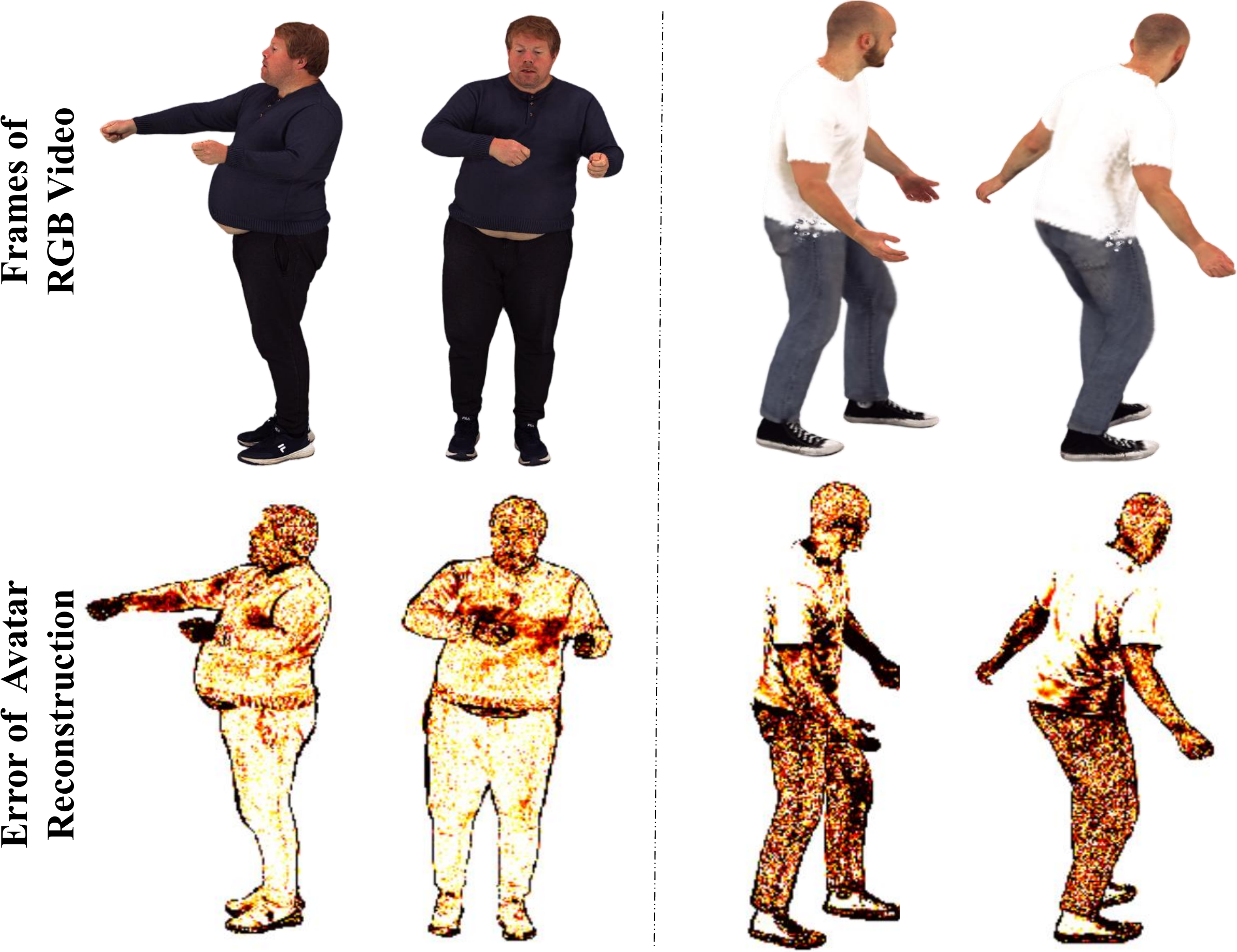}
		\caption{\textbf{Limitation of conventional human gaussian splatting methods.} The second row shows the error between the reconstructed 3D avatar from the corresponding input image in the first row and the 3D ground truth, with darker colors indicating larger reconstruction errors. The hand exhibits the largest reconstruction error across the entire body.}
		\label{fig:intro}
	\end{figure}
	
	Implicit representations bypass grid-based structures by predicting occupancy values over a continuous volumetric space~\cite{NeuMan,vid2avatar}. Although NeRF-based methods offer greater flexibility in modeling, they struggle to accurately capture human pose, facial expressions, and motion due to the absence of explicit structural priors. Subsequent works~\cite{ICON,obja,PAMIR} attempt to address these limitations by incorporating structural priors such as normal maps, the SMPL model, and depth information. However, NeRF-based methods typically incur high computational costs and suffer from slow rendering speeds.
	
	Recently, 3D Gaussian Splatting (3DGS)~\cite{3DGS} has emerged as a powerful technique for efficient, high-fidelity scene reconstruction, inspiring the development of Human Gaussian Splatting methods that leverage 3D Gaussians to model human geometry and appearance. These approaches, such as HUGS~\cite{HUGS} and GaussianAvatar~\cite{hu2024gaussianavatar}, aim to combine the parametric models with the rendering efficiency of 3DGS. X-Avatar~\cite{X-avatar} attempts to improve deformation modeling by assigning individual modules to different body parts. However, these approaches lack the ability to support subtle body parts such as facial and hand details, as illustrated in Figure~\ref{fig:intro}. 
	Although the full-body pose space of SMPL-X~\cite{SMPLX2019} and the hand structure of MANO~\cite{MANO} can be leveraged, standard Gaussian-based avatars still lack the flexibility to model fine non-rigid deformations, particularly for dynamic hand motions and facial expressions. In addition, these methods often suffer from poor temporal consistency, resulting in noticeable flickering under fast motions.
	This limitation hinders their applicability to high-quality human avatar rendering and animation.

\begin{figure*}[htbp]
		\centering
		\includegraphics[width=1\textwidth]{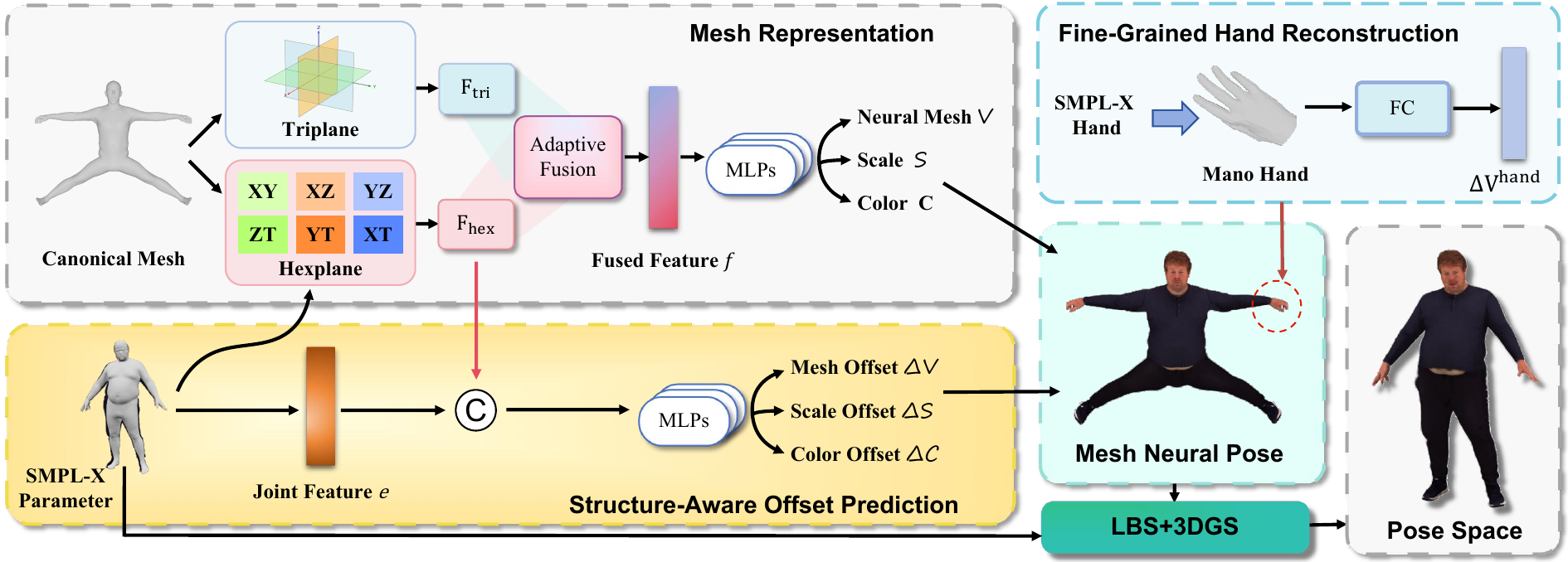}
		\caption{\textbf{Our framework for creating animatable avatars from monocular videos.} We first initialize a set of 3D Gaussians in the canonical space via sampling points from a SMPL-X mesh.The human Gaussians are parameterized by their mean locations and the time $t$ in the canonical space, and adaptive feature fusion is performed using a HexPlane and a triplane to obtain the final representation $f$.Use MLPs to estimate the initial position offset, scale and color.To address the limitations of model-based reconstruction in capturing fine non-rigid deformations, we further introduce a structure-aware offset prediction module that explicitly incorporates joint priors.In addition, we adopt the more detailed MANO model to specifically reconstruct the hands, providing higher fidelity compared to the coarse hand modeling in SMPL-X.Finally, the avatar is animated based on driving signals, SMPL-X parameters, and camera configurations, and is rendered into screen space using 3D Gaussian Splatting (3DGS).}
		\label{fig:SFGS}
	\end{figure*}

	In this work, we propose Structure-aware Fine-grained Gaussian Splatting (SFGS), a 3D Gaussian-based representation for realistic human reconstruction. An overview of our method is illustrated in Figure~\ref{fig:SFGS}.Given SMPL-X parameters and monocular RGB video frames, SFGS leverages the intrinsic joint priors of SMPL-X and a lightweight MLP architecture to effectively correct geometric errors caused by scanned meshes, enabling the reconstruction of more lifelike 3D avatars. To maintain temporal coherence across frames, we introduce a joint Triplane–Hexplane modeling scheme for constructing 3D Gaussian features, which ensures consistent motion dynamics without increasing the training cost. 
	To model the color and position offsets of each Gaussian, we design a structure-aware geometry and color offset module, which enhances the modeling of facial expressions and loose garments within a single-stage training pipeline. 
	Furthermore, to address mismatches between hand poses and captured meshes, we incorporate a residual correction structure that effectively mitigates pose misalignment.
	During inference, SFGS allows frame-wise rendering from arbitrary viewpoints based on any given pose. 
	Extensive experiments on the NeuMan and X-Humans datasets demonstrate both qualitative and quantitative improvements in our approach.
	
	Our contributions are summarized as follows:
	\begin{itemize}
		\item \textbf{Effective structure-coherent avatar reconstruction:} We revisit limitations of existing Human Gaussian Splatting methods and propose SFGS, which enables coherent reconstruction of pose-dependent variations.
		\item \textbf{High-fidelity and fine-grained motion modeling:} SFGS models both hexplane and triplane features, and incorporates structure-aware and fine-grained modules to achieve high-fidelity human motion reconstruction.
	\end{itemize}
	
	\section{Related Work}
	\noindent\textbf{Human Avatar Reconstruction:} Human avatar reconstruction aims to reconstruct a digital representation of a human body from various inputs such as images, videos, or 3D scans. According to the modeling strategy, it can be divided into explicit modeling methods \cite{SCAPE,2007,MGNet} and implicit function modeling methods \cite{LIF,IFN}. Explicit modeling methods involve fitting an explicit parametric human model \cite{SMPLX2019,FLAME} to an image, relying on low-dimensional parameters to represent human pose and shape. However, these methods face challenges in capturing intricate details such as clothing and hair. A common strategy to address this is to introduce pose-dependent offsets to the mesh, enhancing the model’s ability to represent dynamic variations. 
	In contrast, model-free methods overcome these limitations by predicting the occupancy values of a volumetric space. PIFu~\cite{PIFU} exploits a Multi Layer Perceptron (MLP) to model an implicit function that predicts the occupancy value of a given point by leveraging pixel aligned features extracted from the input image. Follow-up works~\cite{ICON, obja, PAMIR} further integrate priors such as normal maps, the SMPL model, and depth information to better handle complex human poses and challenging input conditions. Besides methods utilizing a single-view image as input, multi-view scenarios~\cite{DBLP} offer a richer source of information from different viewpoints, leading to improved reconstruction results. 
	There are also 3D human reconstruction methods~\cite{HumanNeRF,NeuralBody} that represent a 3D human as a NEural Radiance Field (NERF). MonoHuman~\cite{yu2023monohuman} harnesses temporal cues from video to disambiguate the inherent perceptual ambiguity of a single frame, achieving high-fidelity 3D reconstruction. However, NeRF-based methods suffer from slow training and rendering.
	
	
	\noindent\textbf{Human Gaussian Splatting:} Gaussian Splatting~\cite{3DGS} has emerged as a powerful and efficient alternative for 3D reconstruction. HUGS~\cite{HUGS} is a neural human representation based on 3D Gaussians that enables novel pose and view synthesis via a learned forward deformation module in canonical space. GauHuman~\cite{GauHuman} uses articulated Gaussian Splatting with integrated human priors to enable fast training and high-quality rendering. Generalizable human Gaussians regresses 3D Gaussian parameters from the 2D UV space of a human template, enabling photorealistic view synthesis from sparse inputs~\cite{kwon2024generalizable}. 
	Animatable-3D-Gaussian~\cite{animatableGS} introduces a deformation network to transform Gaussians from a canonical space to observation space, along with a time-dependent ambient occlusion module to address dynamic shadows. 
	A drivable human avatar model is proposed to enhance expressiveness from monocular RGB videos by leveraging 3D Gaussians and SMPL-X~\cite{hu2024expressive}. Exavatar~\cite{Exavatar} is an expressive whole-body 3D avatar built from a short monocular video using a hybrid mesh and 3D Gaussian representation.
	Drivable-3D-Gaussian-Avatars~\cite{D3GA} leverages tetrahedral cages and cage-based deformation fields to model the body and individual garments. 
	DAGSM~\cite{DAGSM} tackles disentangled avatar generation with Gaussian Splatting-enhanced meshes. Other works~\cite{HRAvatar,HiFi4G} focus on creating high-quality and compact Gaussian representations for specific body parts. However, these works lack fine-grained modeling, especially for human hands.
	\noindent\textbf{High-Fidelity Avatar Generation:} Generating high-fidelity 3D avatars from limited inputs has been a central challenge. Early approaches rely on parametric 3D morphable models~\cite{3dmm} to reconstruct facial geometry and appearance, but these methods are often constrained by the limited expressiveness of linear shape spaces. The field has since evolved along several interconnected fronts to overcome these limitations. For instance, \cite{liao2023car} aims to generate detailed clothed human avatars from a single image, and \cite{zhang2025high2} reconstructs high-fidelity, lightweight meshes directly from point clouds.
	Beyond holistic reconstruction, a substantial body of research delves into high-fidelity modeling of specific components that are critical for realism. HADES~\cite{Liao_2025_ICCV} models hair with dynamic and explicit strands to achieve unprecedented detail. Recent works~\cite{li2025towards5,expavatar} focus on generating high-fidelity facial avatars for unseen expressions and speech-driven animation, respectively.
	There is also high-fidelity avatar generation from text descriptions~\cite{zhang2025rodinhd,zhang2025avatarstudio}. 
	Most these works aim to generate high-fidelity avatar from text prompts or multimodal inputs. So far, only a few works have explored high-fidelity avatar reconstruction from video based on Gaussian Splatting, but they still fail to capture fine-grained human details.
	
	\section{Preliminaries}
	
	\subsection{Human Gaussian Splatting}
	
	Gaussian Splatting~\cite{3DGS} represents a 3D scene as a collection of anisotropic 3D Gaussians. Each Gaussian $\mathcal{G}_i$ is defined by a mean position $\mathbf{\mu}_i \in \mathbb{R}^3$, a covariance matrix $\Sigma_i \in \mathbb{R}^{3 \times 3}$, an RGB color $\mathbf{c}_i \in \mathbb{R}^3$, and an opacity value $\alpha_i \in [0, 1]$. The final image is generated by projecting these Gaussians onto the image plane and applying alpha-blended splatting in a differentiable rendering pipeline. Formally, the color at pixel $\mathbf{p}$ is rendered as:
	\begin{equation}
		\mathbf{C}(\mathbf{p}) = \sum_{i=1}^N w_i(\mathbf{p}) \cdot \mathbf{c}_i,
	\end{equation}
	where $w_i(\mathbf{p})$ is the weight of the $i$-th Gaussian at pixel $\mathbf{p}$, determined by both its spatial influence and opacity falloff.
	
	To model dynamic humans, recent works~\cite{HUGS,GauHuman,Exavatar} extend Gaussian Splatting by introducing a deformation function $\mathcal{D}_t$, which maps each Gaussian from a canonical space to its posed configuration at frame $t$:
	\begin{equation}
		\mathbf{\mu}_i^{(t)} = \mathcal{D}_t(\mathbf{\mu}_i), \quad \Sigma_i^{(t)} = \mathcal{J}_t \Sigma_i \mathcal{J}_t^\top,
	\end{equation}
	where $\mathcal{J}_t$ is the Jacobian of the deformation at time $t$. This formulation enables a shared set of canonical Gaussians to represent motion over time, significantly reducing memory usage and improving temporal consistency across frames.
	
	\subsection{Incorporating Parametric Human Models}
	
	To guide deformation and introduce pose-awareness, a widely adopted strategy is to incorporate parametric human models like SMPL~\cite{SMPL}, which defines a deformable human mesh $\mathcal{M}$ as follows:
	\begin{equation}
		\mathcal{M}(\boldsymbol{\beta}, \boldsymbol{\theta}) = W(T(\boldsymbol{\beta}, \boldsymbol{\theta}), \boldsymbol{J}(\boldsymbol{\beta}), \boldsymbol{\theta}, \mathcal{W}),
	\end{equation}
	where $\boldsymbol{\beta}$ and $\boldsymbol{\theta}$ represent shape and pose parameters, respectively. The function $T(\cdot)$ generates a template mesh under the given parameters, and $W(\cdot)$ applies linear blend skinning based on joint positions $\boldsymbol{J}$ and skinning weights $\mathcal{W}$. The deformation of each Gaussian point can then be aligned with this skeleton-based motion, ensuring that Gaussians move consistently with the body.
	
	Although SMPL effectively models body movements, it omits fine-grained components such as facial expressions and hand articulation, which are essential for constructing expressive digital humans. To overcome these limitations, SMPL-X~\cite{SMPLX2019}, a more comprehensive body model that extends SMPL with detailed control over the hands, face, and eye gaze, is employed, resulting in a total of 55 articulated joints. Within this framework, SMPL-X parameters serve as pose and expression priors to inform Gaussian deformation, guiding both geometric structure and appearance generation for lifelike and controllable avatars.
	
	\section{Methodology}
	Existing methods of Human Gaussian Splatting~\cite{HUGS,GauHuman,Exavatar} fall short in capturing fine-grained geometry details, and exhibit poor temporal consistency, causing noticeable flickering artifacts or structural instability across video frames. To address these limitations, we present Structure-aware Fine-grained Gaussian Splatting (SFGS) to reconstruct 3D human avartar from a monocular video. 
	
	\subsection{Coherent Mesh Representation}
	Although the SMPL-X model provides a strong structural prior of the human body via low-dimensional pose and shape parameters, its inherently linear formulation limits its ability to represent complex non-rigid deformations. Such deformations, including garment wrinkles and fine motion details, often exhibit strong locality and temporal continuity. To better represent non-rigid dynamics while balancing static and dynamic modeling capabilities, we propose a coherent mesh representation that integrates triplane and hexplane encoding mechanisms atop an upsampled canonical 3D human mesh. 
	
	For the triplane feature, we project each $i$-th point $\mathbf{v}_i$ of the 3D human mesh onto the three canonical planes (XY, YZ, XZ), and bilinearly sample the corresponding feature maps to obtain $\mathbf{f}_{x y}^i, \mathbf{f}_{y z}^i, \mathbf{f}_{x z}^i \in \mathbb{R}^d$. The final triplane representation $\mathbf{f}_{\operatorname{tri}}\left(\mathbf{v}_i\right)$ is obtained by concatenating the sampled features along the three canonical planes:
	\begin{equation}
		\mathbf{f}_{\operatorname{tri}}\left(\mathbf{v}_i\right)=\left[\mathbf{f}_{x y}^i ; \mathbf{f}_{y z}^i ; \mathbf{f}_{x z}^i\right] \in \mathbb{R}^{3 d}.
	\end{equation}
	
	For the hexplane feature, we augment static 3D mesh with timestamp to form $\mathbf{v}_i^t=\left[\mathbf{v}_i, t_i\right] \in \mathbb{R}^4$, and project it onto six 2D planes (XY, XZ, XT, YZ, YT, ZT). The bilinearly sampled features are concatenated as the hexplane feature:
	\begin{equation}
		\mathbf{f}_{\operatorname{hex}}\left(\mathbf{v}_{i}, t_{i}\right)=\left[\mathbf{f}_{x y}^{i} ; \mathbf{f}_{x z}^{i} ; \mathbf{f}_{x t}^{i} ; \mathbf{f}_{y z}^{i} ; \mathbf{f}_{y t}^{i} ; \mathbf{f}_{z t}^{i}\right] \in \mathbb{R}^{6 d}.
	\end{equation}
	
	The hexplane feature is fed into a MLP layer to match the three plane feature distribution and obtain the aligned hexplane feature $\tilde{\mathbf{f}}_{i}^{\mathrm{hex}}\in\mathbb{R}^{3C}$.
	For points on the facial region, we use a dedicated triplane and hexplane with higher resolution and expressive capacity. This allows for capturing fine-grained dynamic expressions and facial details.
	
	Finally, we fuse ${\mathbf{f}}_{i}^{\mathrm{tri}}\in\mathbb{R}^{3C}$ and  $\tilde{\mathbf{f}}_{i}^{\mathrm{hex}}\in\mathbb{R}^{3C}$ through a learned adaptive weight. The final fused feature is given by a weighted combination:
	\begin{equation}\mathbf{f}_{\mathrm{fused}}=(1-\alpha)\odot\mathbf{f}_{\mathrm{tri}}+\alpha\odot\mathbf{f}_{\mathrm{hex}},
	\end{equation}
	where $\odot$ denotes element-wise multiplication, the fusion weight is computed as:
	\begin{equation}
		\alpha=\mathrm{MLP}([\mathbf{f}_\mathrm{tri};\mathbf{f}_\mathrm{hex}]).
	\end{equation}
	This fusion allows the network to adaptively balance static geometry with dynamic temporal changes. The fused feature is fed into MLPs to regress 3D offset, scale, and color.
	
	\subsection{Structure-Aware Offset Prediction}
	The regressed Gaussian properties encode identity- and environment-related information that does not change with pose. To accurately capture pose-dependent geometric deformations, especially in highly articulated regions such as the hands and face, additional pose-aware modeling is required. 
	To this end, we propose a Structure-Aware Offset Prediction module that explicitly leverages skeletal structure to predict per-Gaussian offsets in a pose-conditioned manner. Specifically, we adopt the 55 joints defined by the SMPL-X model—including 22 body joints, 3 facial joints, and 30 hand joints—and assign each Gaussian a dominant joint by taking the $\arg\max$ over the skinning weight matrix. This naturally establishes a structured correspondence between Gaussian points and human skeletal joints.

\begin{figure}[tbp]
		\centering
		\includegraphics[width=0.5\textwidth]{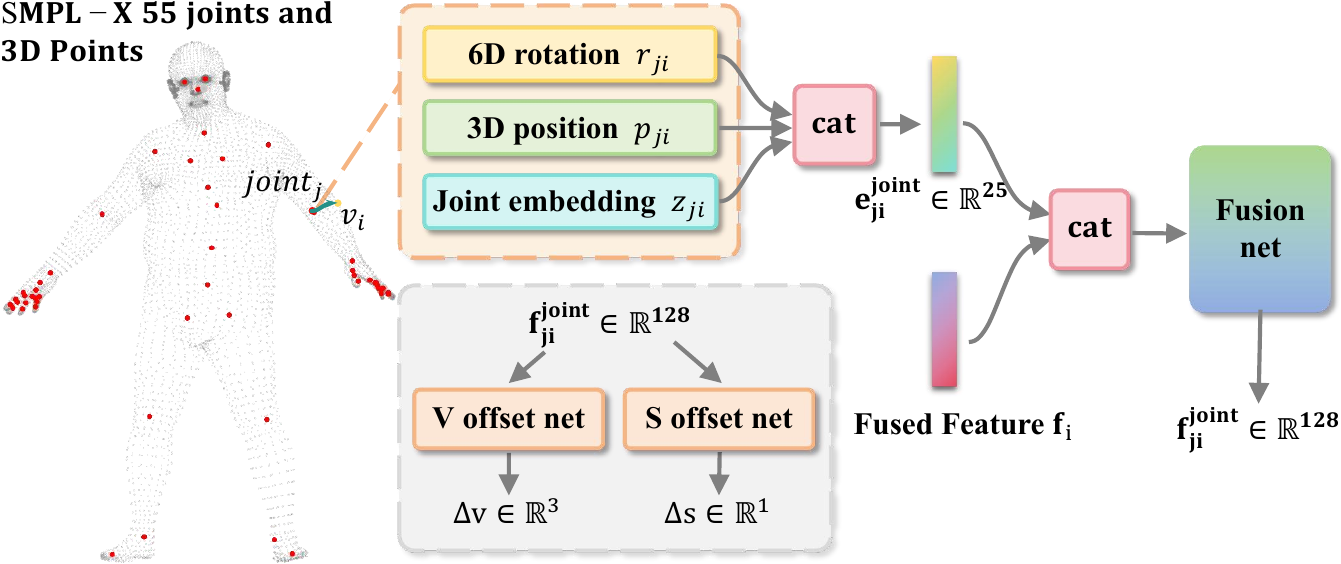}
		\caption{Motivated by the intuition that each Gaussian point should deform under the influence of its most relevant joint, we introduce a joint-aware feature $\mathbf{e}_{\text{joint}}$ to control its scale and positional deformation. A similar mechanism is employed to model the color offset.}
		\label{fig:fig_joint}
	\end{figure}

	Figure~\ref{fig:fig_joint} depicts our structure-aware approach in detail.For each Gaussian point, we extract its fused feature $\mathbf{f}_i^{\mathrm{fused}}$ and construct a joint-aware feature $\mathbf{e}{j_i}$ corresponding to its most influential joint $j_i$. This joint feature is formed by concatenating three components: the 6D joint rotation $\mathbf{r}_{j_i} \in \mathbb{R}^6$, the 3D joint position $\mathbf{p}_{j_i} \in \mathbb{R}^3$, and a learnable joint embedding $\mathbf{z}_{j_i} \in \mathbb{R}^{16}$. The resulting joint feature is:
	\begin{equation}\mathbf{e}_i^{\mathrm{joint}}=[\mathbf{r}_{j_i},\mathbf{p}_{j_i},\mathbf{z}_{j_i}]\in\mathbb{R}^{25}.
	\end{equation}
	We concatenate the fused feature with $\mathbf{e}_i^{\mathrm{joint}}$ and feed it into a lightweight MLP: 
	\begin{equation}
		\mathbf{h}_i=\mathrm{MLP}([\mathbf{f}_i,\mathbf{e}_i^{\mathrm{joint}}]).
	\end{equation}
	The output $\mathbf{h}_i$ is passed through two separate linear branches to predict the Gaussian mean offset $\Delta \boldsymbol{V}_i \in \mathbb{R}^3$ and the scale offset $\Delta S_i \in \mathbb{R}$.
	
	With structural offsets applied to both the Gaussian mean and scale, refined Gaussian parameters are computed as:
	\begin{align}
		\mathbf{V}_{i}=\mathbf{V}_{i}+\Delta\boldsymbol{V}_i, \\
		\mathrm{S}_{i}=\exp(\mathrm{S}_{i}+\Delta S_i),
	\end{align}
	where the exponential function $\exp(\cdot)$ ensures the positivity of the scale and the anisotropic deformations across axes.
	
	This structure-aware design enables the network to learn pose-dependent deformation patterns that are spatially sensitive and respect the human body topology. It enhances the geometric fidelity of highly non-rigid regions such as the hands and face, thereby facilitating more realistic and dynamic reconstructions.
	Unlike SG-GS~\cite{SG-GS}, which requires storing additional semantic labels, or X-Avatar~\cite{X-avatar}, which relies on manual part segmentation, our method avoids introducing any extra semantic priors or region-based grouping. 
	Despite its simplicity, this module captures temporally coherent and fine-grained non-rigid deformations, leading to improved geometry quality overall.
	
	To model pose-dependent color variations, we design a structure-aware color offset module that predicts the RGB offset $\Delta \mathbf{C}_i$ for each Gaussian point. Specifically, we compute a locally fused pose embedding $\mathbf{p}_i$ by aggregating joint rotations$\mathbf{r}j$ weighted by the point’s skinning weights:
	\begin{equation}
		\mathbf{p}_i=\sum_{j=1}^Jw_{i,j}\cdot\mathbf{r}_j.
	\end{equation}
	Unlike the naive baseline, which flattens all joint rotations into a global pose vector and repeats it for every point, our method leverages the skeletal structure to compute point-specific pose features. This allows each point to selectively attend to the joints that influence it most, thus achieving better structure-awareness. In addition, we incorporate the per-point normal vector $\mathbf{n}_i$ to capture local surface orientation, which facilitates more accurate modeling of view-dependent appearance. Finally, the color offset is predicted by an MLP as follows:
	\begin{equation}
		\Delta \mathbf{C}_i = \mathrm{MLP}([\mathbf{f}_i, \mathbf{p}_i, \mathbf{n}_i]).
	\end{equation}
	
	
	\subsection{Fine-Grained Hand Reconstruction}
	Although SMPL-X incorporates hand pose and shape parameters, its relatively coarse mesh topology and global optimization often result in inaccurate hand registration. To address these limitations, we introduce a fine-grained hand reconstruction module for high-fidelity correction. This module is based on the MANO~\cite{MANO}, which specializes in articulated hand modeling. 
	
	Hand surface deformations are represented as a function of pose $\boldsymbol{\theta}_{\text{mano}}$ and shape $\boldsymbol{\beta}_{\text{mano}}$:
	\begin{equation}
		\mathbf{V}^{\text{MANO}} = \mathcal{W}\left(\mathbf{T}(\boldsymbol{\beta}_{\text{mano}}, \boldsymbol{\theta}{\text{mano}}), \boldsymbol{\theta}_{\text{mano}}, \mathcal{J}(\boldsymbol{\beta}_{\text{mano}})\right),
	\end{equation}
	where $\mathbf{V}^{\text{MANO}}$ is calculated on the basis of the shape parameters $\boldsymbol{\beta}_{\text{mano}}$ and pose parameters $\boldsymbol{\theta}_{\text{mano}}$ through the transformation function $\mathcal{W}$ and the joint position function $\mathcal{J}$.
	Training with the residuals of MANO to obtain an independent parameter space specifically for the hands, separate from the body, helps achieve more accurate hand details.
	
	To enable residual transfer, we leverage the fact that the MANO and SMPL-X hand meshes share the same topology. However, despite having a consistent topology, their geometries differ due to the underlying training data. MANO is trained on high-quality hand scans, capturing detailed surface deformations, while SMPL-X’s hand model is integrated into a full-body mesh and lacks the fine-grained details of the hand. We define the initial residual as:
	\begin{equation}
		\Delta \mathbf{V}^{\text{hand}} = \mathcal{M}(\mathbf{V}^{\text{MANO}} - \mathbf{V}^{\text{SMPL-X}}_{\text{hand}}),
	\end{equation}
	where$\Delta \mathbf{V}^{\text{hand}}$ represents the vertex-wise geometric differences between the MANO and SMPL-X hand meshes.
	
	To further incorporate pose-dependent variations, we design a small MLP that refines this residual based on the current frame's hand pose $\mathbf{P}_{\text{hand}}$:
	\begin{equation}
		\Delta\mathbf{V}^{\mathrm{refined}}=\mathrm{MLP} \begin{pmatrix} [\Delta\mathbf{V}^{\mathrm{hand}},\mathbf{P}_{\mathrm{hand}}] \end{pmatrix},
	\end{equation}
	where the MLP learns to adjust the residual according to the hand’s articulation. The refined residuals are then added back to the SMPL-X hand vertices to enhance local hand geometry, capturing finer details such as joint articulation and muscle bulging that are present in the MANO model.

	
	
	
	
	\subsection{Loss Functions}
	We apply Linear Blend Skinning (LBS) to deform the canonical-space vertices into the final posed shape:
	\begin{equation}
		\mathbf{V}_{\mathrm{pose}}=\mathrm{LBS}(\bar{\mathbf{V}},\theta,\mathbf{W}_{\mathrm{pose}}),
	\end{equation}
	where $\theta$ is the pose parameter of SMPL-X and $\mathbf{W}$ represents the skinning weight. 
	
	We then render the posed Gaussians using a differentiable volumetric renderer. Each Gaussian point is parameterized by its position $\mathbf{v}_i \in \mathbf{V_\mathrm{pose}}$, scale $\mathbf{s}_i$, color $\mathbf{c}_i$, and opacity $\alpha_i$. The final rendered image $\hat{\mathbf{I}}$ is obtained via:
	\begin{equation}\hat{\mathbf{I}}=\mathrm{Render}(\left\{\mathbf{v}_i,\mathbf{s}_i,\mathbf{c}_i,\alpha_i\right\}_{i=1}^N),
	\end{equation}
	where $N$ is the number of Gaussians, and $\mathrm{Render}(\cdot)$ denotes the differentiable Gaussian renderer.
	
	Our objective function is designed to jointly optimize the geometry, texture, and structural consistency of the 3D Gaussian avatar. For each training frame, we minimize the following objective:
	\begin{equation}
		\mathcal{L} = \mathcal{L}_{\text{img}} + \mathcal{L}_{\text{face}} + \mathcal{L}_{\text{reg}},
	\end{equation}
	where $\mathcal{L}_{\text{img}}$ supervises the full-body image reconstruction, $\mathcal{L}_{\text{face}}$ enhances appearance consistency in facial regions,
	and $\mathcal{L}_{\text{reg}} $ applies geometry and structure regularizations.
	
	\noindent \textbf{Full-body Image Loss:} To reconstruct photorealistic human appearances, we impose multi-term image supervision on the rendered images:
	\begin{equation}
		\mathcal{L}_{\mathrm{img}}=\lambda_{\mathrm{1}}\cdot(\mathcal{L}_{\mathrm{rgb}}+\mathcal{L}_{\mathrm{ssim}}+\mathcal{L}_{\mathrm{lpips}})+\lambda_{\mathrm{a}}\cdot\mathcal{L}_{\mathrm{lab}}+\lambda_{\mathrm{g}}\cdot\mathcal{L}_{\mathrm{grad}},
	\end{equation}
	where the LPIPS and SSIM losses encourage perceptual fidelity, while RGB and LAB losses constrain low-level color consistency. To encourage sharper rendering and better alignment of local edges, the gradient loss that matches image gradients along both x and y axes is also designed.
	
	\noindent \textbf{Face Loss:} To enhance the realism and consistency of facial textures, we use UV aligned facial textures in the FLAME model as supervision. Use a differentiable renderer to project this texture onto the facial image and monitor the L1 distance between this image and the facial region rendered by our Gaussian model.
	
	\noindent \textbf{Geometry Regularization:} We employ a set of regularizers to ensure stable and plausible geometry:
	\begin{equation}
		\mathcal{L}_{\mathrm{reg}}=\lambda_\mu\cdot\mathcal{L}_{\mathrm{mean}}+\lambda_s\cdot\mathcal{L}_{\mathrm{scale}}+\lambda_{\mathrm{lap}}\cdot\mathcal{L}_{\mathrm{lap}}+\lambda_{\mathrm{joint}}\cdot\mathcal{L}_{\mathrm{joint}},
	\end{equation}
	where $\mathcal{L}_{\mathrm{mean}}$ and $\mathcal{L}_{\mathrm{scale}}$ apply L2 penalties on the Gaussian offsets and scale parameters to maintain local coherence, $\mathcal{L}_{\mathrm{lap}}$ computes over both Gaussian means and scales. 
	The $\mathcal{L}_{\mathrm{lap}}$ encourages local smoothness and reduces spatial noise through a graph structure constructed from adjacent nodes, where higher weights are assigned to artifact-prone regions such as the face and mouth to enhance geometric continuity. Finally, the $\mathcal{L}_{\mathrm{joint}}$ constrains the optimized joint positions with the joints of the original SMPL-X model to prevent attitude drift.
	
	\section{Experiments}
	
	\begin{table}[bp]
		\centering
		\begin{tabular}{lccc}
			\toprule
			Method & PSNR ↑ & SSIM ↑ & LPIPS ↓ \\
			\midrule
			NeuMan~\cite{NeuMan}        &  29.32 &  0.972 & 0.014 \\
			Vid2Avatar~\cite{vid2avatar}    &  30.70 & 0.980 &  0.014 \\
			GaussianAvatar~\cite{hu2024gaussianavatar}     & 29.94 & 0.980 & 0.012 \\
			3DGS-Avatar~\cite{3DGS}    &  28.99 &  0.974 &  0.016 \\
			Exavatar~\cite{Exavatar}    & 34.80 &  0.984 & 0.009 \\
			\midrule
			SFGS (ours)      & \textbf{35.34} & \textbf{0.985} & \textbf{0.009} \\
			\bottomrule
		\end{tabular}
		\caption{\textbf{Quantitative comparison on the Neuman dataset.} To align with prior work, we report the mean results over the sequences bike, citron, jogging, and seattle.}
		\label{tab:Neuman_all}
	\end{table}
	
	\begin{table*}[tbp]
		\centering
		\resizebox{\textwidth}{!}{
			\begin{tabular}{l|ccc|ccc|ccc|ccc|ccc|ccc}
				\hline
				\textbf{Methods} &
				\multicolumn{3}{c|}{\textbf{Seattle}} &
				\multicolumn{3}{c|}{\textbf{Citron}} &
				\multicolumn{3}{c|}{\textbf{Parking}} &
				\multicolumn{3}{c|}{\textbf{Bike}} &
				\multicolumn{3}{c|}{\textbf{Jogging}} &
				\multicolumn{3}{c}{\textbf{Lab}} \\
				& PSNR$\uparrow$ & SSIM$\uparrow$ & LPIPS$\downarrow$ 
				& PSNR$\uparrow$ & SSIM$\uparrow$ & LPIPS$\downarrow$
				& PSNR$\uparrow$ & SSIM$\uparrow$ & LPIPS$\downarrow$
				& PSNR$\uparrow$ & SSIM$\uparrow$ & LPIPS$\downarrow$
				& PSNR$\uparrow$ & SSIM$\uparrow$ & LPIPS$\downarrow$
				& PSNR$\uparrow$ & SSIM$\uparrow$ & LPIPS$\downarrow$ \\
				\hline
				Vid2Avatar~\cite{vid2avatar} 
				& 16.90 & 0.51 & 0.27 
				& 15.96 & 0.59 & 0.28 
				& 18.51 & 0.65 & 0.26 
				& 12.44 & 0.39 & 0.54 
				& 16.36 & 0.46 & 0.30 
				& 15.99 & 0.62 & 0.34 \\
				
				NeuMan~\cite{NeuMan} 
				& 18.42 & 0.58 & 0.20 
				& 18.39 & 0.64 & 0.19 
				& 17.66 & 0.66 & 0.24 
				& 19.05 & 0.66 & 0.21 
				& 17.57 & 0.54 & 0.29 
				& 18.76 & 0.73 & 0.23 \\
				
				HUGS~\cite{HUGS} 
				& 19.06 & 0.67 & 0.15 
				& 19.16 & 0.71 & 0.16 
				& 19.44 & 0.73 & 0.17 
				& 19.48 & 0.67 & 0.18 
				& 17.45 & 0.59 & 0.27 
				& 18.79 & 0.76 & 0.18 \\
				
				ExAvatar~\cite{Exavatar} 
				& 36.98 & 0.991 & 0.006 
				& 35.43 & 0.987 & 0.008 
				& 37.16 & 0.988 & 0.012 
				& 32.81 & 0.981 & 0.011 
				& 32.72 & 0.977 & 0.013 
				& 34.69 & 0.981 & 0.013 \\
				
				SFGS (ours) 
				& \textbf{37.58} & \textbf{0.991} & \textbf{0.006} 
				& \textbf{35.49} & \textbf{0.986} & \textbf{0.009} 
				& \textbf{37.50} & \textbf{0.988} & \textbf{0.010} 
				& \textbf{34.75} & \textbf{0.983} & \textbf{0.012} 
				& \textbf{33.55} & \textbf{0.978} & \textbf{0.013} 
				& \textbf{34.85} & \textbf{0.987} & \textbf{0.011} \\
				\hline
			\end{tabular}
		}
		\caption{\textbf{Quantitative comparison on six sequences of the Neuman dataset}. The results of Exavatar~\cite{Exavatar} are implemented by ourselves, while the results of other models are taken from \cite{HUGS}.
		}
		\label{tab:neuman6}
	\end{table*}

	\begin{table*}[tbp]
		\centering
		\resizebox{0.85\textwidth}{!}{ 
			\begin{tabular}{l|ccc|ccc|ccc}
				\toprule
				\textbf{Methods} & \multicolumn{3}{c|}{\textbf{00028}} & \multicolumn{3}{c|}{\textbf{00034}} & \multicolumn{3}{c}{\textbf{00087}} \\
				& PSNR$\uparrow$ & SSIM$\uparrow$ & LPIPS$\downarrow$ 
				& PSNR$\uparrow$ & SSIM$\uparrow$ & LPIPS$\downarrow$ 
				& PSNR$\uparrow$ & SSIM$\uparrow$ & LPIPS$\downarrow$ \\
				\midrule
				X-Avatar~\cite{X-avatar} 
				& 28.57 & 0.976 & 0.026 
				& 28.05 & 0.965 & 0.035 
				& 30.89 & 0.970 & 0.030 \\
				ExAvatar~\cite{Exavatar}
				& 30.58 & 0.981 & 0.018 
				& 28.75 & 0.966 & 0.029 
				& 32.01 & 0.972 & 0.025 \\
				SFGS (ours) 
				& \textbf{31.12} & \textbf{0.983} & \textbf{0.018} 
				& \textbf{29.10} & \textbf{0.967} & \textbf{0.035} 
				& \textbf{32.10} & \textbf{0.972} & \textbf{0.025} \\
				\bottomrule
			\end{tabular}
		}
		\caption{~\textbf{Quantitative comparison on subjects \texttt{00028}, \texttt{00034}, and \texttt{00087} on the X-Humans dataset}. Higher PSNR/SSIM and lower LPIPS indicate better performance.}
		\label{tab:tab1}
	\end{table*}
	
	\begin{table*}[t]
		\centering
		\small
		\renewcommand{\arraystretch}{1.2}
		\setlength{\tabcolsep}{6pt}
		\resizebox{\textwidth}{!}{
			\begin{tabular}{ll|ccc|ccc|ccc|ccc}
				\toprule
				\multirow{2}{*}{\textbf{Subset}} & \multirow{2}{*}{\textbf{Method}} & \multicolumn{3}{c|}{~\textbf{CD $\downarrow$}} & \multicolumn{3}{c|}{~\textbf{CD-MAX $\downarrow$}} & \multicolumn{3}{c|}{~\textbf{NC $\uparrow$} }& \multicolumn{3}{c}{~\textbf{IoU $\uparrow$} }\\
				\cmidrule(lr){3-5} \cmidrule(lr){6-8} \cmidrule(lr){9-11} \cmidrule(lr){12-14}
				& & All & Face & Hand & All & Face & Hand & All & Face & Hand & All & Face & Hand \\
				\midrule
				\multirow{2}{*}{00028} & ExAvatar & 9.86 & 6.30 & 7.30 & 82.13 & ~\textbf{22.29} & 22.43 & 0.916 & ~\textbf{0.837} & 0.932 & 0.924 & 0.751 & 0.854 \\
				& Ours    & ~\textbf{9.06} & ~\textbf{5.88} & ~\textbf{7.11} & ~\textbf{79.39} & 22.77 & ~\textbf{22.04} & ~\textbf{0.927} & 0.831 & ~\textbf{0.934} & ~\textbf{0.939} & ~\textbf{0.785} & ~\textbf{0.870} \\
				\midrule
				\multirow{2}{*}{00034} & ExAvatar & 9.34 & ~\textbf{5.32} & 6.93 & 68.58 & 21.22 & ~\textbf{26.03} & 0.900 &  0.823 & 0.941 &  ~\textbf{0.946} & 0.771 & 0.905 \\
				& Ours    & ~\textbf{9.12} & 5.60 & ~\textbf{6.46} & ~\textbf{67.91} & ~\textbf{20.47} & 26.62 & ~\textbf{0.912} &  ~\textbf{0.825} & ~\textbf{0.965} & 0.945 & ~\textbf{0.825} & ~\textbf{0.923} \\
				\midrule
				\multirow{2}{*}{00087} & ExAvatar & 8.85 & 4.43 & 6.18 & 66.64 & 19.90 & 23.65 & 0.886 & ~\textbf{0.819} & 0.931 & ~\textbf{0.949} & 0.746 & ~\textbf{0.901} \\
				& Ours    & ~\textbf{8.81} & ~\textbf{4.13} & 6.18 & ~\textbf{64.71} & ~\textbf{19.55} & ~\textbf{23.34} & ~\textbf{0.906} & 0.818 & ~\textbf{0.936} & 0.946 & ~\textbf{0.781} & 0.900 \\
				\bottomrule
			\end{tabular}
		}
		\caption{~\textbf{More detailed quantitative results.} The results verify that our method can produce more detailed and high-quality human reconstructions.}
		\label{tab:region_metrics}
	\end{table*}
	
	\begin{figure*}[tbp]
		\centering
		\includegraphics[width=1\textwidth]{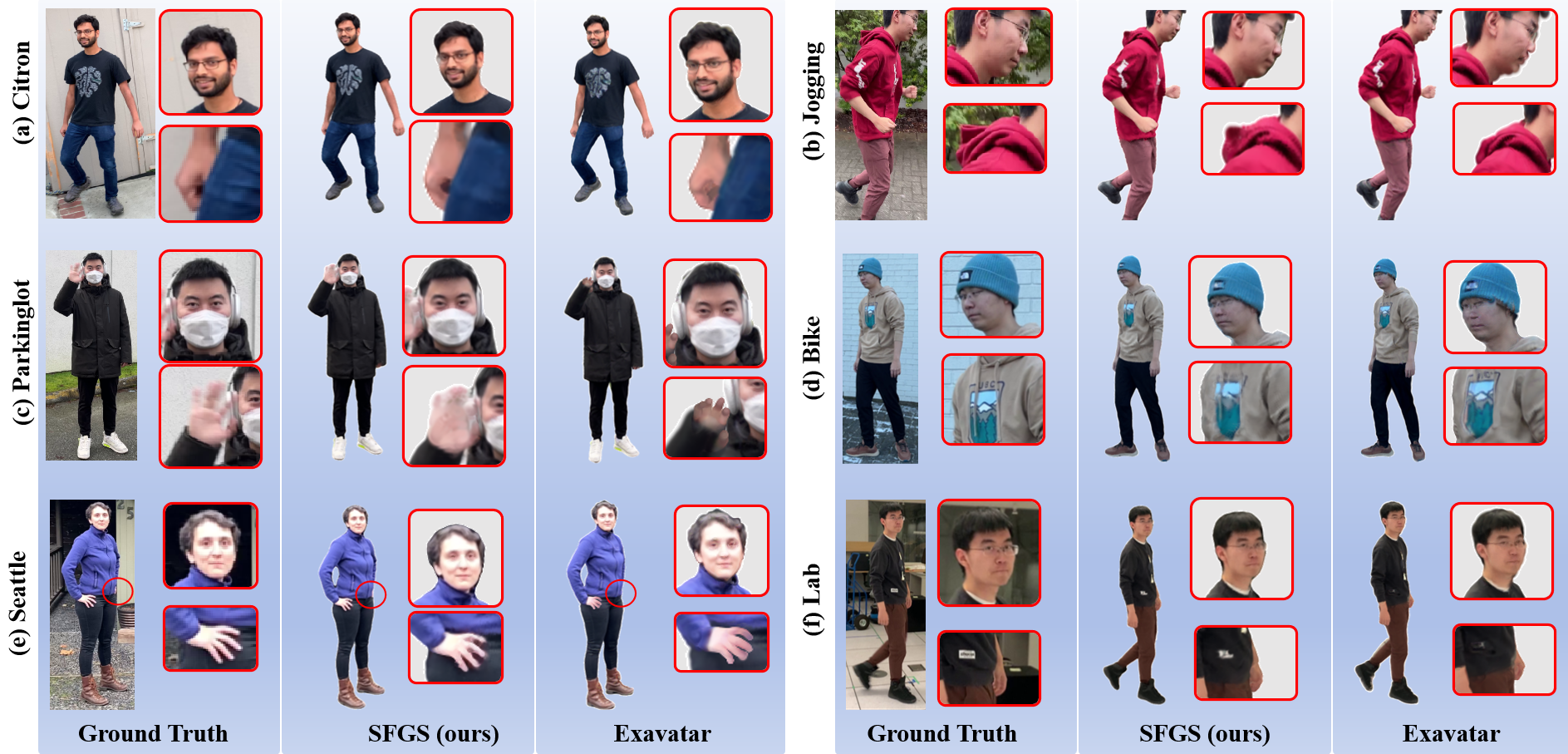}
		\caption{\textbf{Quality comparison on the Neuman dataset.} For each example numbered from (a) to (f), the results are shown from left to right as Ground Truth, SFGS (ours), and ExAvatar~\cite{Exavatar}.}
		\label{fig:fig_neuman}
	\end{figure*}
	
	\begin{figure*}[tbp]
		\centering
		\includegraphics[width=1\textwidth]{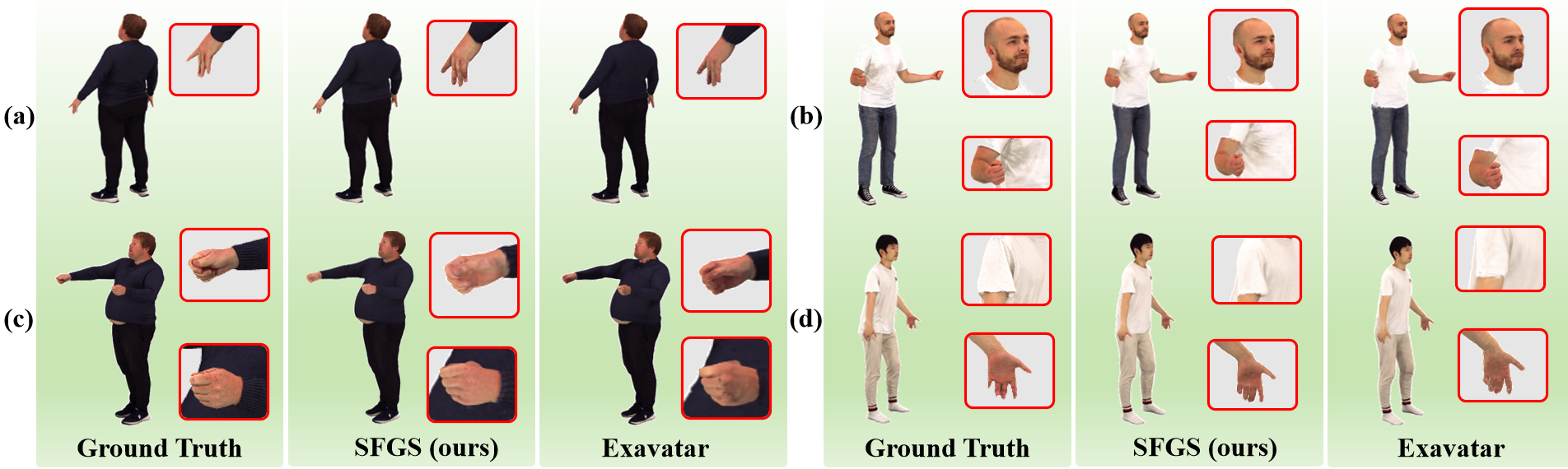}
		\caption{\textbf{Qualitative Comparison on X-humans.} From left to right: Ground Truth,Ours and ExAvatar.}
		\label{fig:fig_comp}
	\end{figure*}
	
	\subsection{Datasets and Implementation Details}
	\noindent\textbf{NeuMan Dataset:} The NeuMan dataset~\cite{NeuMan} consists of six monocular video sequences, each capturing a single subject with a handheld mobile phone, with video durations ranging from 10 to 20 seconds. This dataset presents more diverse apparel and significantly richer, continuous human motions, providing a stronger showcase for the fidelity of our 3D reconstruction. Drawing on previous research~\cite{NeuMan,HUGS}, we use videos of bike, citron, jogging, lab, parking lot and Seattle, which show most areas of the human body.
	Following the official protocol, every sequence is divided into 80 training, 10 validation, and 10 test frames. Evaluation considers only the foreground subject and background regions are excluded from all metrics.
	
	\noindent\textbf{X-Humans Dataset:} The X-Humans dataset~\cite{X-avatar} consists of high-quality 3D textured scans and RGB-D video sequences of multiple human subjects captured in a controlled studio environment. X-Humans offers greater diversity in facial expressions and hand poses, making it more suitable for evaluating expressive and fine-grained human avatars. For fair and consistent evaluation, we use subjects 00028, 00034, and 00087 of different body shapes and clothing styles from the pre-trained models publicly released under the RGB-D protocol.  
	
	
	\noindent\textbf{Implementation Details: }
	In our implementation, we adopt two distinct planar representations, Triplane and Hexplane, to model spatial features and ensure temporal consistency for Gaussian feature extraction. Specifically, the Triplane is represented as $\mathbf{T} \in \mathbb{R}^{3 \times C \times H \times W}$, and the Hexplane is a learnable tensor $\mathbf{H} \in \mathbb{R}^{6 \times C \times H \times W}$, where $C=32$ and $H=W=128$ denote the channel dimension and spatial resolution, respectively. 
	
	We employ MaskedRNN~\cite{NeuMan,HUGS} for subject-background segmentation and sample approximately 170,000 points on the SMPL-X mesh surface. All Multi-Layer Perceptrons (MLPs) are 2-layer networks with a hidden dimension of 256 and ReLU activations. The loss weights are configured as $\lambda_{1} = 0.8$, and $\lambda_{\text{a}} = \lambda_{\text{g}} = 0.03$. The entire framework is trained on a single NVIDIA RTX 4090 GPU.
	
	
	\subsection{Comparison with State-of-the-Arts}
	
	\noindent\textbf{Evaluation on Neuman:} 
	Table~\ref{tab:Neuman_all} shows the average performance on bike, citron, jogging, and Seattle on the Neuman dataset. Our method achieves consistent improvements in PSNR and SSIM while reducing LPIPS, significantly outperforming the state-of-the-arts across all scenarios.
	
	Table~\ref{tab:neuman6} compares our method with other state-of-the-arts on all six datasets. As can be seen, while LPIPS remains unchanged, our PSNR improves by an average of 2.4\% across six scenarios compared with previous methods, with the Bike scenario showing a particularly notable 5.9\% improvement over the previous state-of the-arts.
	
	\noindent\textbf{Evaluation on X-Humans:} Table~\ref{tab:tab1} shows the evaluations on the X-Humans dataset. The results demonstrate that our proposed SFGS achieves superior performance over previous state-of-the-arts across all metrics on different subsets. As shown in Table~\ref{tab:region_metrics}, our method consistently outperforms the baselines across all evaluation metrics and body regions. Notably, for the hand region, SFGS achieves the highest accuracy, reducing the average per-sample Chamfer Distance (CD) by 0.192 ($3.8\% \downarrow$) and improving the IoU by 0.03 ($3.79\% \uparrow$) compared to ExAvatar. This quantitative gain highlights our method's effectiveness in capturing fine-grained geometric details and non-rigid deformations.
	
	
	\begin{figure}[htbp]
		\centering
		\includegraphics[width=0.48\textwidth]{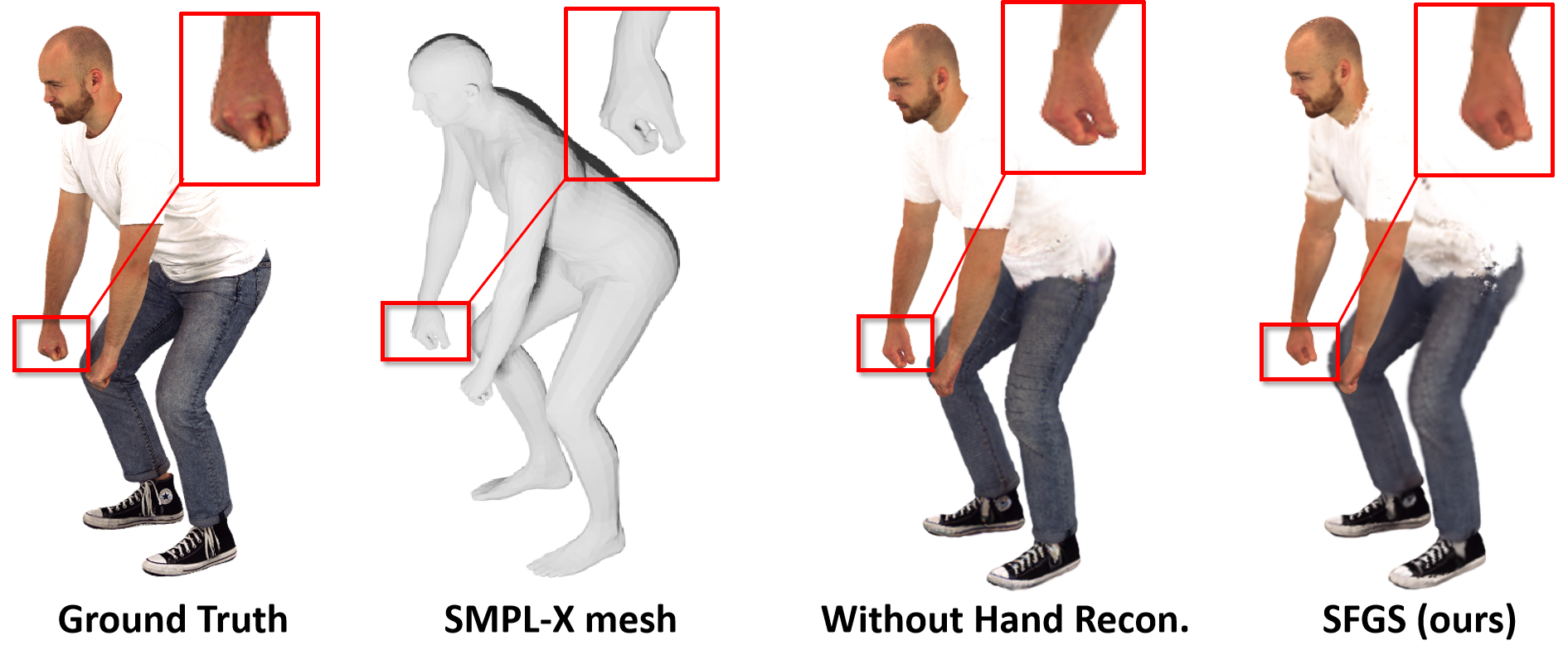}
		\caption{\textbf{Visualizations of the effect of fine-grained hand reconstruction.} The SMPL-X mesh shows noticeable artifacts in the hands, which are mitigated by our approach.}
		\label{fig:handcomp}
	\end{figure}
	
	\begin{figure}[htbp]
		\centering
		\includegraphics[width=0.47\textwidth]{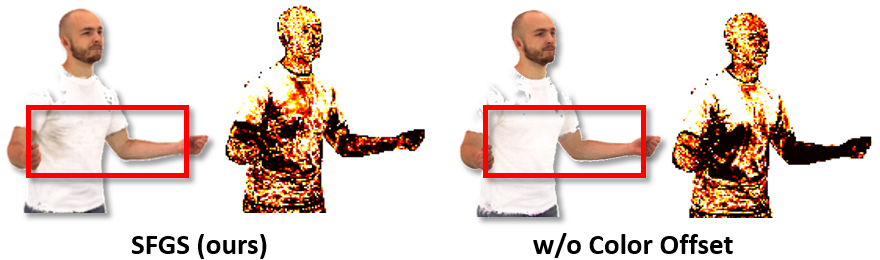}
		\caption{\textbf{Visualizations of the effect of color offset in the structure-aware offset prediction module.} The color offset makes the appearance of wrinkles and skin more consistent with real photos. For each approach, the PSNR heat map is on the right.}
		\label{fig:psnr}
	\end{figure}
	
	\begin{figure}[htbp]
		\centering
		\includegraphics[width=0.47\textwidth]{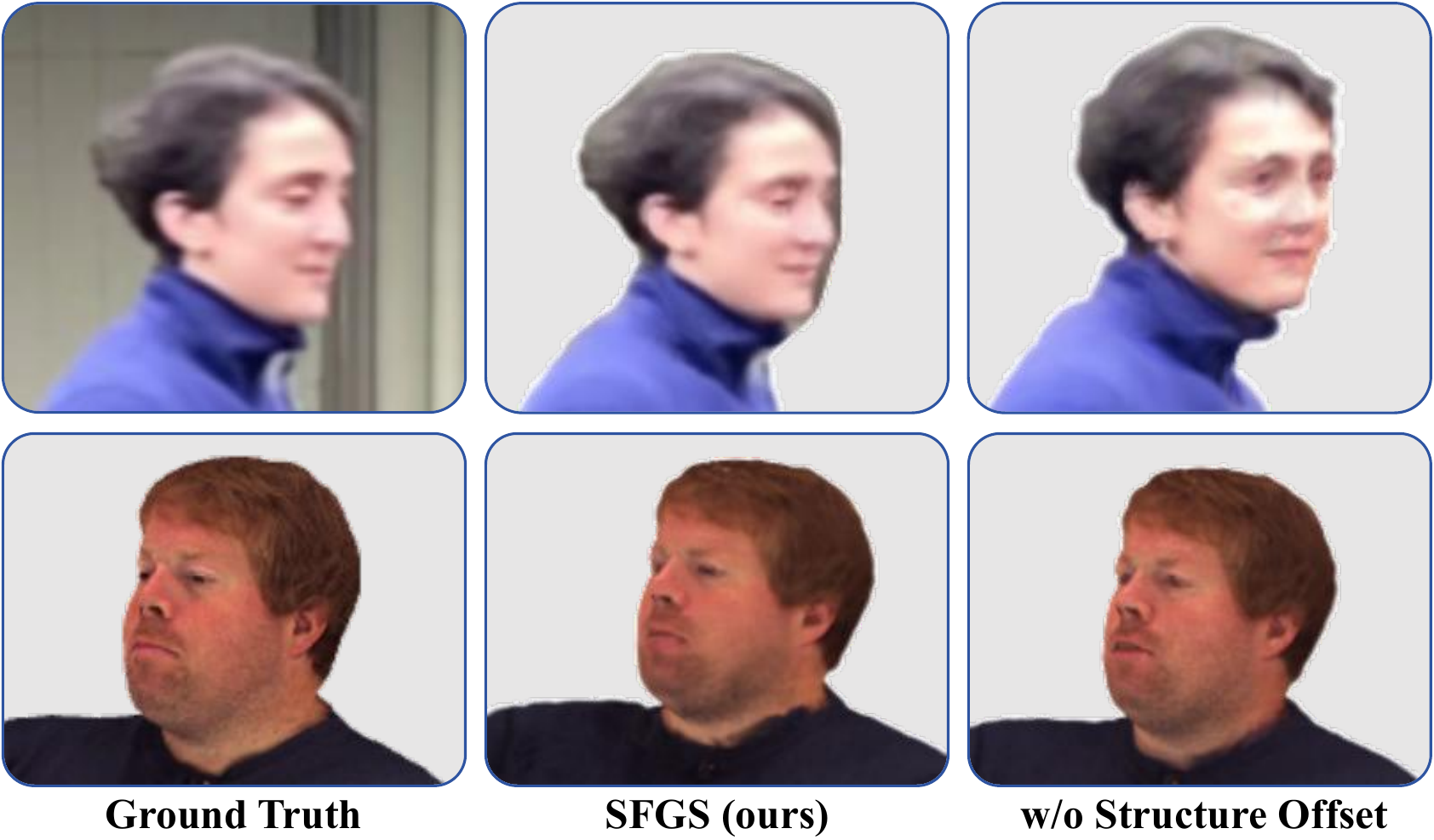}
		\caption{\textbf{Visualizations of the effect of structure offset in the structure-aware offset prediction module.} The structure offset leads to more accurate mouth pose details.} 
		\label{fig:mouth}
	\end{figure}
	
	\noindent\textbf{Visualizations of Avatar Reconstruction: } 
	Figure~\ref{fig:fig_neuman} visualizes rendered images of the reconstructed avatar and compares our method with ExAvatar~\cite{Exavatar}. Our renderings achieve photorealistic results under novel viewpoints and poses. In example (d), our method produces significantly clearer hands and faces. Examples (c) and (e) demonstrate that our renderings preserve hand details much better than previous avatars under novel viewpoints and motions. Furthermore, ExAvatar ignores or shrinks very small body parts, such as the hands in examples (e) and (f). In contrast, our method reincorporates these parts into the rendered result, yielding a closer approximation to the original photo.
	
	As shown in Figure~\ref{fig:fig_comp}, These improvements are further validated on the X-Humans dataset through our Fine-Grained Hand Reconstruction module. As shown in panels (a) and (c), our method accurately captures subtle hand details, such as color consistency and cast shadows, surpassing the baseline in both color fidelity and shadow realism.Overall, our method yields a much closer approximation to the original photos across diverse human subjects and clothing styles.
	
	\begin{table}[tbp]
		\centering
		\resizebox{0.48\textwidth}{!}{
			\begin{tabular}{lccc}
				\toprule
				Method & PSNR ↑ & SSIM ↑ & LPIPS ↓ \\
				\midrule
				Ours w/o Color Offset     & 30.83 & 0.983 & 0.019 \\
				Ours w/o Hand Recon.    & 30.92 & 0.927 & 0.019 \\
				Ours w/o HexPlane         & 31.03 & 0.983 & 0.019 \\
				Ours w/o Structure Offset    & 30.58 & 0.982 & 0.02 \\
				\midrule
				SFGS (ours)  & \textbf{31.12} & \textbf{0.983} & \textbf{0.018} \\
				\bottomrule
		\end{tabular}}
		\caption{\textbf{Results of ablation studies of the proposed SFGS.} Removing any single component results in performance drops, especially for the PSNR metric which evaluates fine details.}
		\label{tab:ablation}
	\end{table}
	
	\subsection{Ablation Studies and Analysis}
	
	The proposed SFGS consists of three modules: Coherent Mesh Representation, Structure-Aware Offset Prediction, and Fine-Grained Hand Reconstruction. The first module is based on HexPlane modeling, while the second predicts both structure and color offsets. Results of ablation studies on the X-Humans dataset are provided in Table~\ref{tab:ablation}.

	\noindent\textbf{Impact of Hand Reconstruction: }Fine-Grained Hand Reconstruction, despite only targeting the hands, contributes a 0.06 improvement in SSIM for the rendered images.
	
	Figure~\ref{fig:handcomp} demonstrates the effect of fine-grained hand modeling, which could help correct the errors introduced by the SMPL-X model during the scanning process.
	
	\noindent\textbf{Impact of HexPlane Modeling: }
	Compared with the variant without HexPlane, HexPlane modeling improves the PSNR and LPIPS metrics, which confirms that HexPlane modeling enhances the coherence of mesh representation.
	
	\noindent\textbf{Impact of Color Offset: }Similarly, applying the color offset yields a 0.29 PSNR improvement, owing to leveraging joint information to guide the color prediction.
	
	Figure~\ref{fig:psnr} further demonstrates the advanced performance of our structure-aware color offset module. It can be seen that the color of the skin and wrinkles is corrected, which reduces the PSNR of the rendering results. 
	
	\noindent\textbf{Impact of Structure Offset: }Incorporating structure-aware offset prediction into geometric structures results in a 0.54 increase in PSNR and a decrease in LPIPS, indicating improved visual fidelity.
	
	Figure~\ref{fig:mouth} demonstrates the superiority of our structure-aware geometric reconstruction method. Compared with offset prediction methods that ignore structural cues, our structure-aware geometry module can focus on the granularity of ultra-detailed features such as the eyes and mouth, making the overall rendering closer to photorealistic results. 
	
	\begin{figure}[tbp]
		\centering
		\includegraphics[width=0.47\textwidth]{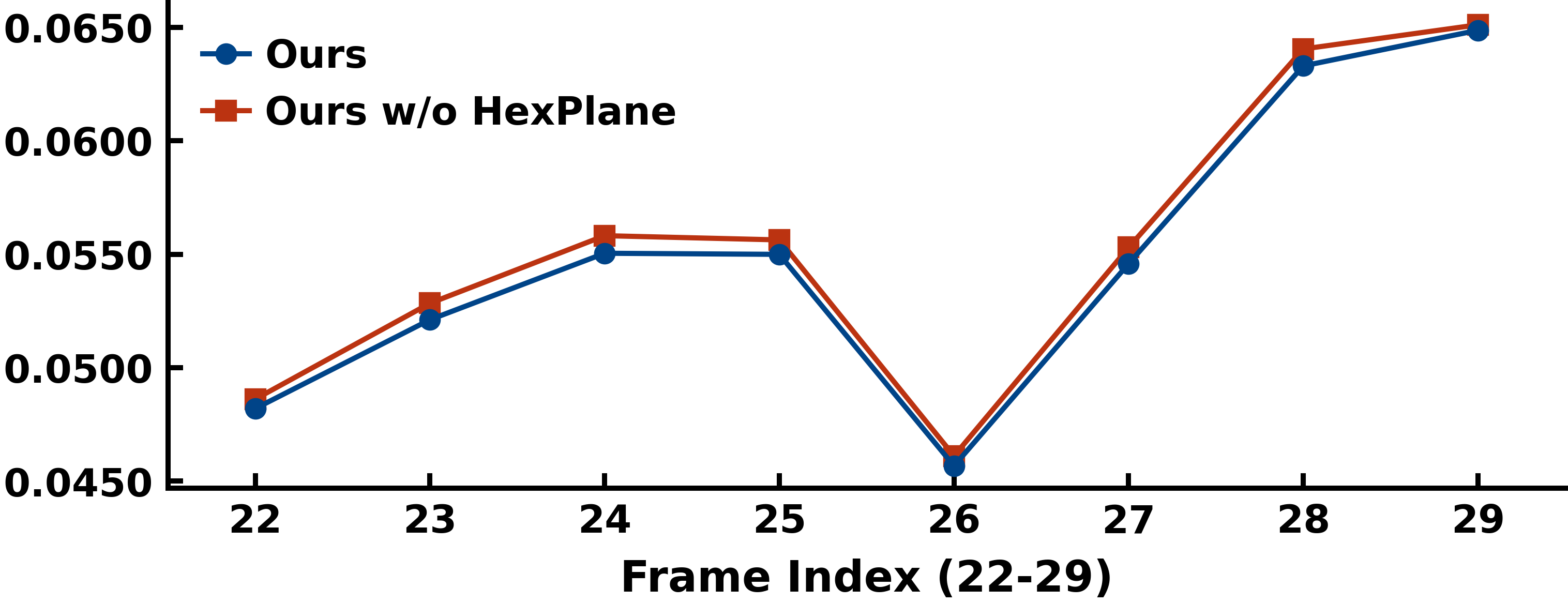}
		\caption{\textbf{Temporal consistency comparison with and without HexPlane. }The y-axis denotes the tc-LPIPS metric evaluated on 7 consecutive frames. Models equipped with HexPlane exhibit consistently lower LPIPS values, suggesting enhanced temporal coherence.}
		\label{fig:hexplane}
	\end{figure}
	
	\begin{table}[tbp]
		\centering
		\resizebox{0.48\textwidth}{!}{
			\begin{tabular}{lccc}
				\toprule
				Hyperparameters & PSNR ↑ & SSIM ↑ & LPIPS ↓ \\
				\midrule
				\textbf{$\lambda_{\mathrm{a}} = 0.3$}, \textbf{$\lambda_{\mathrm{g}} = 0.03$}     & 30.83 & 0.968 & 0.02 \\
				\textbf{$\lambda_{\mathrm{a}} = 0.3$}, \textbf{$\lambda_{\mathrm{g}} = 0.005$}    & 30.95 & 0.976 & 0.19 \\
				\textbf{$\lambda_{\mathrm{a}} = 0.5$}, \textbf{$\lambda_{\mathrm{g}} = 0.01$}         & 30.65 & 0.982 & 0.02 \\
				\textbf{$\lambda_{\mathrm{a}} = 0.1$}, \textbf{$\lambda_{\mathrm{g}} = 0.01$}    & 30.93 & 0.975 & 0.02 \\
				\midrule
				\textbf{$\lambda_{\mathrm{a}} = 0.3$}, \textbf{$\lambda_{\mathrm{g}} = 0.01$}  & \textbf{31.12} & \textbf{0.983} & \textbf{0.018} \\
				\bottomrule
		\end{tabular}}
		\caption{\textbf{Sensitivity analysis of hyperparameters.}}
		\label{tab:sana}
	\end{table}
	
	\noindent\textbf{Impact of HexPlane Modeling: }
	We compute the temporal consistency LPIPS (tc-LPIPS) to assess whether visual "flickering" occurs between adjacent frames when the camera and human eye move synchronously. As illustrated in Figure~\ref{fig:hexplane}, the consecutive rendering results demonstrate that introducing Hexplane significantly reduces tc-LPIPS, indicating its ability to suppress flickering and jitter, thereby producing temporally smoother dynamic details.
	
	\noindent\textbf{Impact of Hyperparameter Configuration: }
	\textbf{$\lambda_{\mathrm{g}}$} controls the strength of this loss, enhancing image edge sharpness and preserving local details without introducing noise. Through sensitivity analysis, we determined the optimal hyperparameters for the Lab Loss and Gradient Loss components. As shown in Table ~\ref{tab:sana}, the best-performing configuration was found to be \textbf{$\lambda_{\mathrm{a}} = 0.3$} and \textbf{$\lambda_{\mathrm{g}} = 0.01$}.
	
	\noindent\textbf{Running Time Comparison: } 
	As shown in Table \ref{tab:fps_comparison}, our SFGS method achieves a peak rendering speed of 30 FPS, surpassing all baseline approaches, including the highly competitive Exavatar (26 FPS). On an NVIDIA RTX 4090, our method facilitates real-time performance at $800 \times 1200$ resolution. We further analyzed the structure-aware offset prediction module by substituting the MLP with an attention mechanism. This variation, however, led to a substantial decline in rendering speed and a PSNR drop to 30.88. Therefore, we retained the MLP structure to ensure high-fidelity rendering without compromising real-time performance.
	\begin{table}[tbp]
		\centering
		\caption{Comparison of rendering speed (FPS) with state-of-the-art 3D human avatar generation methods. }
		\label{tab:fps_comparison}
		\begin{tabular}{lc}
			\toprule
			Method & FPS $\uparrow$ \\
			\midrule
			Neuman      & 0.02 \\
			Vid2avatar  & 0.04 \\
			X-avatar    & 0.12 \\
			Exavatar    & 26.00 \\
			\textbf{SFGS (Ours)} & \textbf{30.00} \\
			\bottomrule
		\end{tabular}
	\end{table}
	
	\section{Conclusion}
	SFGS addresses key challenges in human avatar modeling by leveraging structure-aware modules for geometry and appearance, enabling accurate reconstruction of fine-grained hand gestures and expressive facial motions. By combining SMPL-X and MANO with Gaussian-based representation, our method achieves anatomically consistent, temporally coherent animations across complex articulations.Despite these advancements, our method lacks adaptiveness when handling wide or bulky body regions, where insufficient upsampled point density may result in blurred rendering boundaries. In future work, we aim to enhance the generalization ability to diverse human body types and clothing styles, and simplify the motion editing process by incorporating sketch- or text-guided interaction for more intuitive avatar control.
	
	{
		\small
		\bibliographystyle{ieeenat}
		\bibliography{main}
	}
	
	
\end{document}